\documentclass[]{article}
\usepackage{proceed2e}
\usepackage{mathtools}
\usepackage{times}
\usepackage{natbib}
\usepackage{graphicx}
\usepackage{caption}
\usepackage{subcaption} 
\usepackage{hyperref}
 \usepackage{amssymb}
\usepackage{amsmath}
\usepackage{bbm}
\newcommand{\E}{\mathbb{E}}
\newcommand{\Ss}{\mathcal{S}}

\DeclareMathOperator*{\argmax}{arg\,max}

\usepackage[algo2e]{algorithm2e}
\usepackage{algorithm}
\usepackage{bm}
\newenvironment{nscenter}
 {\parskip=3pt\par\nopagebreak\centering}
 {\par\noindent\ignorespacesafterend}

 \newenvironment{itemize*}%
  {\vspace{-2.5\topsep} \begin{itemize}%
    \setlength{\parskip}{0pt}
  \setlength{\itemsep}{0.75pt plus 1pt}}%
  {\end{itemize} \vspace{-2.5\topsep}}

\title{A Reinforcement Learning Approach to Weaning of Mechanical Ventilation in Intensive Care Units}

\title{A Reinforcement Learning Approach to Weaning of Mechanical Ventilation in Intensive Care Units}

\author{{\bf Niranjani Prasad} \\
Princeton University \\
\And
{\bf Li-Fang Cheng}  \\
Princeton  University \\
\And
{\bf Corey Chivers}   \\
Penn Medicine \\
\And
{\bf Michael Draugelis}   \\
Penn Medicine \\
\And
{\bf Barbara E. Engelhardt}   \\
Princeton  University    \\
}

\begin{document}

\maketitle

\begin{abstract}
The management of invasive mechanical ventilation, and the regulation of sedation and analgesia during ventilation, constitutes a major part of the care of patients admitted to intensive care units. Both prolonged dependence on mechanical ventilation and premature extubation are associated with increased risk of complications and higher hospital costs, but clinical opinion on the best protocol for weaning patients off of a ventilator varies. 
This work aims to develop a decision support tool that uses available patient information to predict time-to-extubation readiness and to recommend a personalized regime of sedation dosage and ventilator support. To this end, we use off-policy reinforcement learning algorithms to determine the best action at a given patient state from sub-optimal historical ICU data. We compare treatment policies from fitted Q-iteration with extremely randomized trees and with feedforward neural networks, and demonstrate that the policies learnt show promise in recommending weaning protocols with improved outcomes, in terms of minimizing rates of reintubation and regulating physiological stability.
%\looseness=-1

\end{abstract}

\section{Introduction}
\label{submission}

Mechanical ventilation is one of the most widely used interventions in admissions to the intensive care unit (ICU): around 40\% of patients in the ICU are supported on invasive mechanical ventilation at any given hour, accounting for 12\% of total hospital costs in the United States (\cite{ambrosino,wunsch2013icu}). These are typically patients with acute respiratory failure or compromised lung function caused by some underlying condition such as pneumonia, sepsis, or heart disease, or cases in which breathing support is necessitated by neurological disorders, impaired consciousness, or weakness following major surgery. As advances in healthcare enable more patients to survive critical illness or surgery, the need for mechanical ventilation during recovery has risen. %\looseness=-1

Closely coupled with ventilation in the care of these patients is sedation and analgesia, which are crucial to maintaining physiological stability and controlling pain levels of patients while intubated.
The underlying condition of the patient, as well as factors such as obesity or genetic variations, can have a significant effect on the pharmacology of drugs, and cause high inter-patient variability in response to a given sedative (\cite{patel2012sedation}), lending motivation to a personalized approach to sedation strategies. %\looseness=-1

\textit{Weaning} refers to the process of liberating patients from mechanical ventilation. The primary diagnostic tests for determining whether a patient is ready to be extubated involve screening for resolution of the underlying disease, haemodynamic stability, assessment of current ventilator settings and level of consciousness, and finally a series of spontaneous breathing trials (SBTs). Prolonged ventilation---and corresponding over-sedation---is associated with post-extubation delirium, drug dependence, ventilator-induced pneumonia, and higher patient mortality rates (\cite{hughes2012sedation}), in addition to inflating costs and straining hospital resources. Physicians are often conservative in recognizing patient suitability for extubation, however, as failed breathing trials or premature extubations that necessitate reintubation within 48-72 hours can cause severe patient discomfort and result in even longer ICU stays (\cite{krinsley2012optimal}). Efficient weaning of sedation and ventilation is therefore a priority both for improving patient outcomes and reducing costs, but a lack of comprehensive evidence and the variability in outcomes between individuals and subpopulations means there is little agreement in clinical literature on the best weaning protocol (\cite{conti2014sedation,Goldstone986}). %\looseness=-1

In this work, we aim to develop a decision support tool that leverages available patient information in the data-rich ICU setting to alert clinicians when a patient is ready for initiation of weaning, and to recommend a personalized treatment protocol. We explore the use of off-policy reinforcement learning algorithms, namely fitted Q-iteration (FQI) with different regressors, to determine the optimal treatment at each patient state from sub-optimal historical patient treatment profiles. 
The setting fits naturally into the framework of reinforcement learning as it is fundamentally a sequential decision making problem rather than purely a prediction task: we wish to choose the best possible action at each time---in terms of sedation drug and dosage, ventilator settings, initiation of a spontaneous breathing trial, or extubation---while capturing the stochasticity of the underlying process, the delayed effects of actions, and the uncertainty in state transitions and outcomes.

The problem poses a number of key challenges: there are a multitude of factors that can potentially influence patient readiness for extubation, including some not directly observed in ICU chart data, such as a patient's inability to protect their airway due to muscle weakness. The data that is recorded is often sparse and noisy. In addition, there is potentially an extremely large space of possible sedatives and ventilator settings that can be leveraged during weaning. We are also posed with the problem of interval censoring, as in other intervention data: given past treatment and vitals trajectories, observing a successful extubation at time $t$ provides us only with an upper bound on the true time to extubation readiness, $t_e\leq t$; on the other hand, if a breathing trial was unsuccessful, there is uncertainty how premature the intervention was. This presents difficulties both when learning the policy and in evaluating policies.

The rest of the paper is organized as follows: Section 2 explores recent efforts in the use of reinforcement learning in clinical settings. In Section 3, we describe the data and methods used here, and Section 4 presents the results. Finally, conclusions and possible directions for further work are discussed in Section 5. %\looseness=-1

\section{Related Work} 
 
The widespread adoption of electronic health records (EHRs) paved the way for a data-driven approach to healthcare, and recent years have seen a number of efforts towards personalized, dynamic treatment regimes.
Reinforcement learning in particular has been explored in various settings, from determining the sequence of drugs to be administered in HIV therapy or cancer treatment, to management of anaemia in haemodialysis patients, and insulin regulation in diabetics. These efforts are typically based on estimating the \emph{value}, in terms of clinical outcomes, of different treatment decisions given the state of the patient.

For example, \cite{ernst2006clinical} applied fitted Q-iteration with a tree-based ensemble method to learn the optimal HIV treatment in the form of structured treatment interruption strategies, in which patients are cycled on and off drug therapy. The observed reward here is defined in terms of the equilibrium point between healthy and unhealthy blood cells in the patient. 
\cite{nonsmallcell} used Q-learning to learn optimal individualized treatment regimens for nonsmall cell lung cancer. The objective is to choose the optimal first and second lines of therapy and the optimal initiation time for the second line treatment such that the overall survival time is maximized. The Q-function with time-indexed parameters is approximated using a modification of support vector regression (SVR) that explicitly handles right-censored data. In this setting, right-censoring arises in measuring the time of death from start of therapy: given that a patient is still alive at the time of the last follow-up, we merely have a lower bound on the exact survival time. %\looseness=-1

\cite{EscandellMontero201447} compared the performance of both Q-learning and fitted Q-iteration with current clinical protocol for informing the delivery of erythropoeisis-stimulating agents (ESAs) for treating anaemia. The drug administration strategy is modeled as a Markov decision process (MDP), with the state space expressed by current values and change in haemoglobin levels, the most recent ESA dosage, and the patient subpopulation. The action space is a set of four discretized ESA levels, and the reward function is designed to maintain haemoglobin levels within a healthy range while avoiding abrupt changes. %\looseness=-1

On the problem of administering anaesthesia in an ICU setting, \citet{moore2004intelligent} applied Q-learning with eligibility traces to the administration of intravenous propofol, modeling patient dynamics according to an established pharmacokinetic model, with the aim of maintaining some level of sedation or consciousness. \cite{Padmanabhan} also used Q-learning, for the regulation of both sedation level and arterial pressure as an indicator of physiological stability, using propofol infusion rate. All of the aforementioned work rely on model-based approaches to reinforcement learning, and develop treatment policies on simulated patient data. More recently however, \cite{nemati} consider the problem of heparin dosing to maintain blood coagulation levels within some well-defined therapeutic range, modeling the task as a partially observable MDP, using a dynamic Bayesian network trained on real ICU data, and learning a dosing policy with neural fitted Q-iteration (NFQ). \looseness=-1

There exists some literature on machine learning methods for the problem of ventilator weaning: \cite{mueller2013can} and \cite{kuo2015improvement} look at prediction of weaning outcomes using supervised learning methods, and suggest that classifiers based on neural networks, logistic regression, or naive Bayes, trained on patient ventilator and blood gas data, show promise in predicting successful extubation. \cite{gao2017incorporating} develops association rule networks for naive Bayes classifiers, to analyze the discriminative power of different feature categories toward each decision outcome class, in order to help inform clinical decision making. 
Our paper is novel in its use of reinforcement learning methods to directly tackle policy recommendation for ventilation weaning. Specifically, we incorporate a larger number of possible predictors of weaning readiness, in a 32-dimensional patient state representation, compared with previous works which typically limit features for classification to at most a couple of key vital signs. Moreover, we make use of current clinical protocols to inform the design and tuning of a reward function. %\looseness=-1

\section{Methods}

\subsection{Critical Care Data}

\begin{figure*}[ht]
\centering
\captionsetup{justification=centering}
    \includegraphics[width=0.95\textwidth]{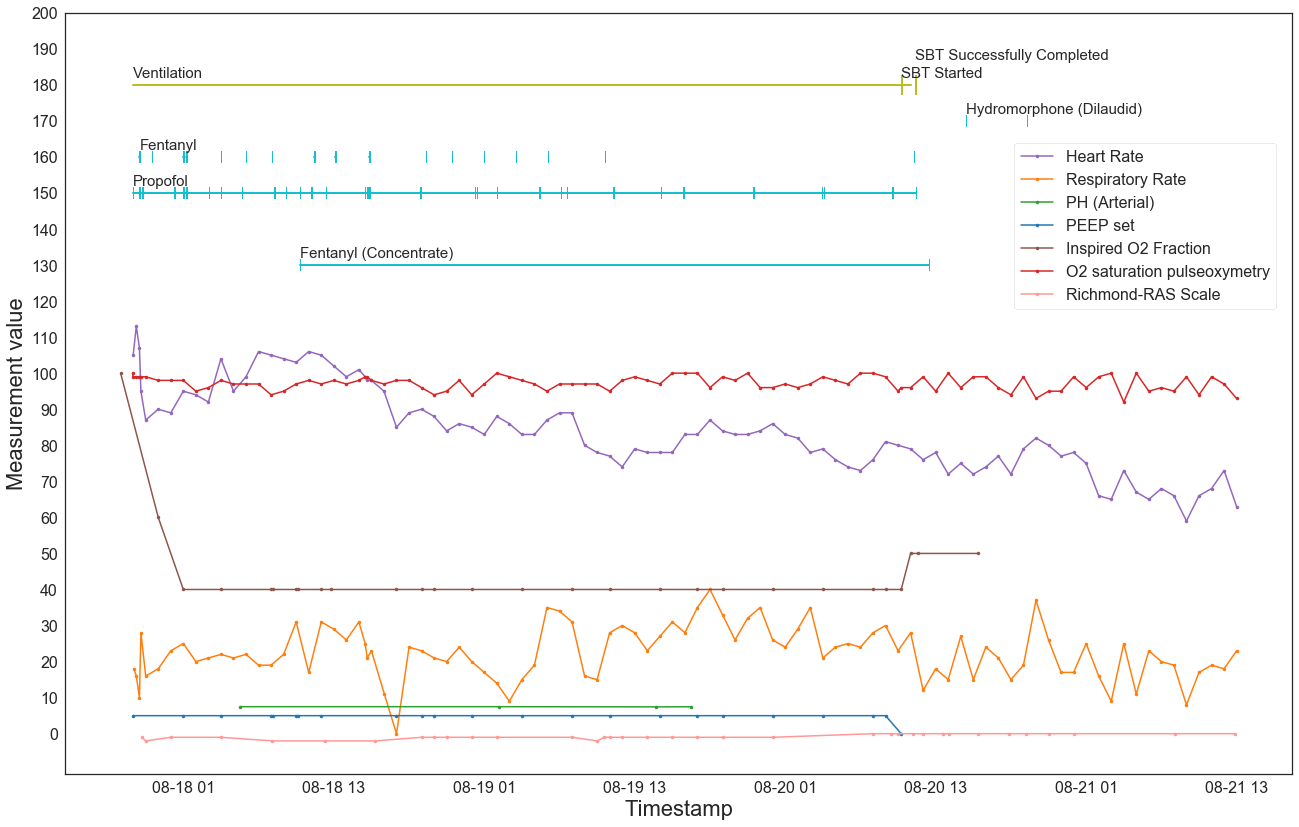}
    \caption{Example ventilated ICU patient. Vitals are measured at a range of sampling intervals. Ventilation times are marked, and multiple administered sedatives (both as continuous IV drips and discrete boli) are shown.}
    \label{fig:ptTraj}
\end{figure*}

We use the Multi-parameter Intelligent Monitoring in Intensive Care (MIMIC III) database (\cite{johnson2016mimic}), a freely available source of de-identified critical care data for 53,423 adult admissions and 7,870 neonates. The data includes patient demographics, time-stamped measurements from bedside monitoring of vitals, administration of fluids and medications, results of laboratory tests, observations and notes charted by care providers, as well as diagnoses, procedures and prescriptions for billing. %\looseness=-1

We extract from this database a set of 8,860 admissions from 8,182 unique adult patients undergoing invasive ventilation. In order to train and test our weaning policy, we filter further to include only those admissions in which the patient was kept under ventilator support for more than 24 hours. This allows us to exclude the majority of episodes of routine ventilation following surgery, which are at minimal risk of adverse extubation outcomes.
We also filter out admissions in which the patient in not successfully discharged from the hospital by the end of the admission, as in cases where the patient expires in the ICU, failure to discharge is largely due to factors beyond the scope of ventilator weaning, and again, a more informed weaning policy is unlikely to have a significant influence on outcomes.
\textit{Failure} in our problem setting is instead defined as prolonged ventilation, administration of unsuccessful spontaneous breathing trials, or reintubation within the same admission---all of which are associated with adverse outcomes for the patient. A typical patient timeline is illustrated in Figure \ref{fig:ptTraj}.
%\looseness=-1

Preliminary guidelines for the weaning protocol, in terms of the desired ranges of physiological parameters (heart rate, respiratory rate, and arterial pH) as well as criteria at time of extubation for the inspired $O_2$ fraction ($FiO_2$), oxygenation pulse oxymetry ($SpO_2$), and positive end-expiratory pressure (PEEP) set, were obtained from clinicians at the Hospital of University of Pennsylvania, HUP (Table \ref{table:penn}). These are used in shaping rewards in our MDP to facilitate learning of the optimal policy. %\looseness=-1

\begin{table}[ht]
\centering
\renewcommand{\arraystretch}{1.2}
\begin{tabular}{| c | c |}
  \hline		
  \textbf{\,\,\,Physiological Stability\,\,\,} & \textbf{\,\,\, Oxygenation Criteria\,\,\,} \\
  \hline\hline
  Respiratory Rate $\leq 30$ &  PEEP (cm $H_2O$) $ \leq 8$ \\
  Heart Rate $\leq 130$ & $SpO_2$ (\%) $\geq 88$ \\
  Arterial pH $\geq 7.3$ & Inspired $O_2$ (\%) $ \leq 50$\\
  \hline
\end{tabular}
\vspace{5pt}
    \caption{Current extubation guidelines at HUP.}
    \label{table:penn}
\vspace{-5pt}
\end{table}

\begin{figure*}[ht]
\centering
\captionsetup{justification=centering}
\begin{subfigure}{0.32\textwidth}
\centering
    \includegraphics[width=\textwidth]{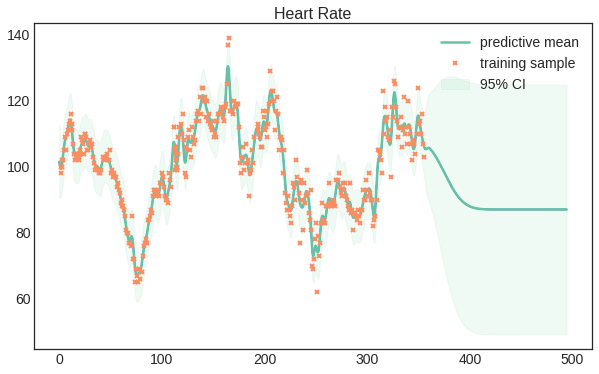}
    %\caption{Heart Rate}
    \label{1}
\end{subfigure}
\begin{subfigure}{0.32\textwidth}
\centering
    \includegraphics[width=\textwidth]{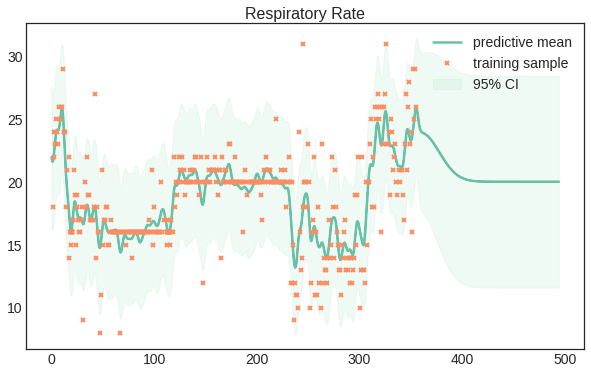}
    %\caption{Arterial pH}
    \label{2}
\end{subfigure}
\begin{subfigure}{0.32\textwidth}
\centering
        \includegraphics[width=\textwidth]{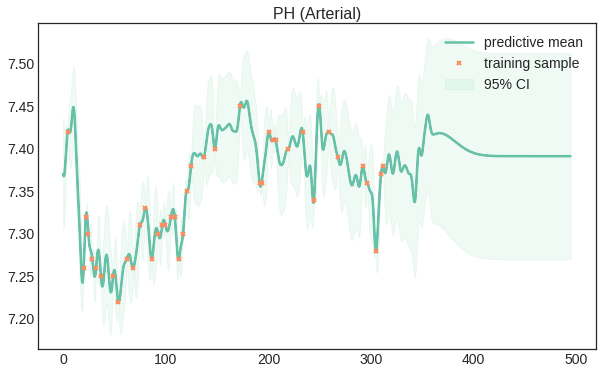}
    %    \caption{Inspired $O_2$ fraction}
        \label{3}
    \end{subfigure}
    \captionsetup{justification=centering}
\begin{subfigure}{0.32\textwidth}
\centering
    \includegraphics[width=\textwidth]{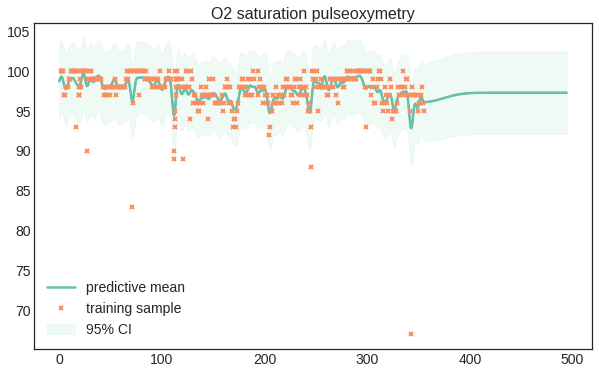}
  %  \caption{Heart Rate}
    \label{4}
\end{subfigure}
\begin{subfigure}{0.32\textwidth}
\centering
    \includegraphics[width=\textwidth]{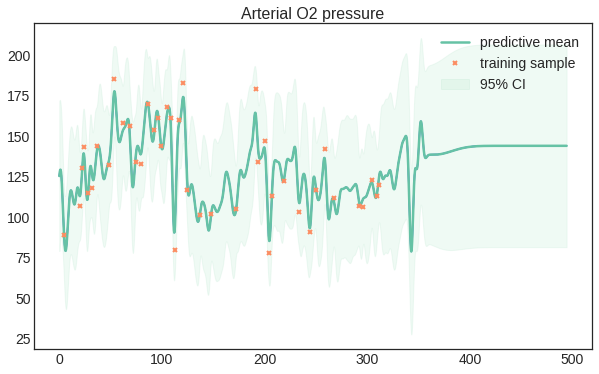}
  %  \caption{Arterial pH}
    \label{5}
\end{subfigure}
\begin{subfigure}{0.32\textwidth}
\centering
        \includegraphics[width=\textwidth]{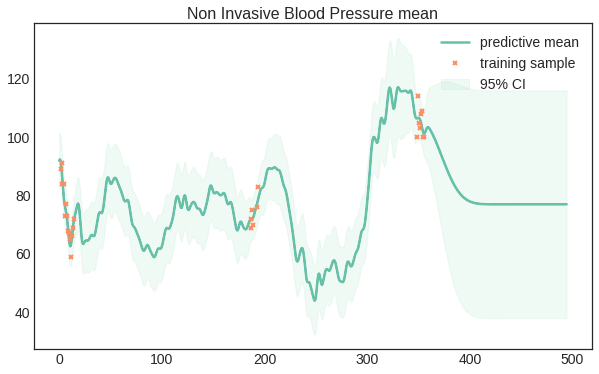}
   %     \caption{Inspired $O_2$ fraction}
        \label{6}
    \end{subfigure}
    \caption{Example trajectories of six vital signs for a single admission, following imputation using Gaussian processes. \\ Twelve vital signs are jointly modeled by the GP.}
     \label{fig:gp_interp}
\end{figure*}

\subsubsection{Preprocessing using Gaussian Processes}

Measurements of vitals and lab results in the ICU data can be irregular, sparse, and error-prone.  Non-invasive measurements such as heart rate or respiratory rate are taken several times an hour, while tests for arterial pH or oxygen pressure, which involve more distress to the patient, may only be administered every few hours as needed. This wide discrepancy in measurement frequency is typically handled by resampling with means in hourly intervals (when we have multiple measurements within an hour), and using sample-and-hold interpolation to impute subsequent missing values. However, patient state---and therefore the need to update management of sedation or ventilation---can change within the space of an hour, and naive methods for interpolation are unlikely to provide the necessary accuracy at higher temporal resolutions.
We therefore explore methods for the imputation of patient state that can enable more precise policy estimation. 

One commonly used approach to resolve missing and irregularly sampled time series data is Gaussian processes (GPs, \citep{stegle2008gaussian, Durichen2015MTGP,ghassemi2015multivariate}).Denoting the observations of the vital signs by $\mathbf{v}$ and the measurement time $\mathbf{t}$, we model
\[
\mathbf{v} = f(\mathbf{t}) + \bm{\varepsilon},
\]
where $\bm{\varepsilon}$ vector represents i.i.d Gaussian noise, and $f(\mathbf{t})$ is the latent noise-free function we would like to estimate. We put a GP prior on the latent function $f(\mathbf{t})$:
\[
f(\mathbf{t}) \sim \mathcal{GP}(m(\mathbf{t}), \kappa(\mathbf{t}, \mathbf{t}')),
\]
where $m(\mathbf{t})$ is the \emph{mean function} and $\kappa(\mathbf{t}, \mathbf{t}')$ is the \emph{covariance function} or \emph{kernel}, which shapes the temporal properties of $f(\mathbf{t})$. 
%For instance, with the squared exponential (SE) kernel, parameterized as
%\[
%\kappa(\mathbf{x}, \mathbf{x}') = \sigma^{2}\exp{\left( -\frac{||\mathbf{x}-%\mathbf{x}'||^{2}}{2\ell^{2}} \right)},
%\]
%the function values are smooth over the input domain. 
%Most previous work explores the function of each physiological signal separately using univariate GP; the temporal correlations between each signal is assumed to be zero. This may cause lost of information (e.g. reported correlation between heart rate and blood pressure~\cite{Nemati2012cardio}) and limit the accuracy of interpolation when the sampling density varies greatly across these physiological signals. Multivariate GPs have shown improvements over univariate GP in medical time series for imputation and forecasting~\cite{Durichen2015MTGP, ghassemi2015multivariate}.
In this work, we use a multi-output GP to account for temporal correlations between physiological signals during interpolation. 
We adapt the framework in \cite{Cheng2017arXiv} to impute the physiological signals jointly by estimating covariance structures between them, excluding the sparse prior settings. We set $m(\mathbf{t}) = \mathbf{0}$ without loss of generality (\cite{Rasmussen2006gpml}), and $\kappa(\mathbf{t}, \mathbf{t}')$ as the kernel in the linear model of coregionalization with the spectral kernel as the basis kernel, allowing us to model both smooth correlations in time and periodic variations of these vital signs and lab results. The full joint kernel for each patient $i$ is defined as: %\looseness=-1
\[
\kappa_{i}(\mathbf{t}_{i}, \mathbf{t}_{i}') = \displaystyle \sum_{q=1}^{Q}{\bm{B}_{q} \otimes \kappa_{q}(\mathbf{t}_{i,*}, \mathbf{t}_{i,*}')},
\]
where $\mathbf{t}_{i,*}$ represents the time vector of each vital sign. Note that this is a simplified representation based on the assumption that we have the same input time vector for each signal, which does not hold in our irregularly sampled data. In practice we have to compute each sub-block $\kappa_{q}(\mathbf{t}_{i,d}, \mathbf{t}_{i,d'}')$ given any pair of input time $\mathbf{t}_{i,d}$ and $ \mathbf{t}_{i,d'}'$ from two signals, indexing by $d$ and $d'$. We use $Q$ to denote the number of mixture kernels, and $\bm{B}_{q}$ to encode the scale covariance between any pair of signals, written as
\[
\begin{array}{cl}
\bm{B}_{q} &= 
\begin{bmatrix}
b_{q, (1, 1)} & b_{q, (1, 2)} & \cdots & b_{q, (1, D)}\\
b_{q, (1, 1)} & \vdots & \ddots & \vdots\\
\vdots & \vdots & \ddots & \vdots \\
b_{q, (D, 1)} & b_{q, (D, 2)} & \cdots & b_{q, (D, D)}\\
\end{bmatrix}
\in \mathbb{R}^{D \times D}.
\end{array}
\] % note: remember to include package bm
The basis kernel is parameterized as
\[
\kappa_{q}(\mathbf{t}, \mathbf{t}') = \exp{(-2\pi^2\tau^{2} v_{q})}\cos{(2\pi\tau\mu_{q})},
\]
\[\
\tau = |\mathbf{t}-\mathbf{t}'|.
\]
We set $Q=2$ and $R=5$ for modeling 12 selected physiological signals ($D=12$) jointly. For each patient, one structured GP kernel is estimated using the implementation in \cite{Cheng2017arXiv}. We then impute the time series with the estimated posterior mean given all the observations across all chosen physiological signals for that patient. 
In choosing the 12 signals, we exclude vitals that take discrete values, such as ventilator mode or the RASS sedation scale; for these, we simply resample with means and apply sample-and-hold interpolation.
After preprocessing, we obtain complete data for each patient, at a temporal resolution of 10 minutes, from admission time to discharge time (Figure~\ref{fig:gp_interp}). 

\subsection{MDP Formulation}

A Markov decision process is defined by 

\begin{itemize*}
\item[(i)] A finite \textbf{state space $\Ss$} such that at each time $t$, the environment (here, the patient) is in state $s_t \in \Ss$.

\item[(ii)] An \textbf{action space $\mathcal{A}$}: at each time $t$, the agent takes action $a_t \in \mathcal{A}$, which influences the next state, $s_{t+1}$.

\item[(iii)] A \textbf{transition function} $P(s_{t+1} | s_t, a_t)$, the probability of the next state given the current state and action, which defines the (unknown) dynamics of the system.

\item[(iv)] A \textbf{reward function} $r(s_t, a_t) \in \mathbb{R}$, the observed feedback following a transition at each time step $t$.

\end{itemize*}

The goal of the reinforcement learning agent is to learn a \textbf{policy}, i.e. a mapping $\pi(s) \rightarrow a$ from states to actions, that maximizes the expected accumulated reward
\[
R^\pi(s_t) = \lim\limits_{T\rightarrow \infty} \E_{s_{t+1}|s_{t,\pi(s_t)}}\sum\limits^T_{t=1}\gamma^tr(s_t,a_t)
\]
over time horizon $T$. The discount factor, $\gamma$, determines the relative weight of immediate and long-term rewards.

Patient response to sedation and readiness for extubation can depend on a number of different factors, from demographic characteristics, pre-existing conditions, and comorbidities to specific time-varying vital signs, and there is considerable variability in clinical opinion on the extent of the influence of different factors. Here, in defining each patient state within an MDP, we look to incorporate as many reliable and frequently monitored features as possible, and allow the algorithm to determine the relevant features.
The state at each time $t$ is a 32-dimensional feature vector that includes fixed demographic information (patient \emph{age, weight, gender, admit type, ethnicity}), as well as relevant physiological measurements, ventilator settings, level of consciousness (given by the Richmond Agitation Sedation Scale, or RASS), current dosages of different sedatives, time into ventilation, and the number of intubations so far in the admission. For simplicity, categorical variables \emph{admit type} and \emph{ethnicity} are binarized as emergency/non-emergency and white/non-white, respectively. %\looseness=-1

In designing the action space, we develop an approximate mapping of six commonly used sedatives into a single dosage scale, and choose to discretize this scale to four different levels of sedation. The action $a_t \in \mathcal{A}$ at each time step is chosen from a finite two-dimensional set of eight actions, where $a_t[0] \in \{0,1\}$ indicates having the patient off or on the ventilator, respectively, and $a_t[1] \in \{0,1,2,3\}$ corresponds to the level of sedation to be administered over the next 10-minute interval:
\[ 
\mathcal{A} = \left\{ 
\begin{bmatrix} 0 \\ 0 \end{bmatrix},
\begin{bmatrix} 0 \\ 1 \end{bmatrix},
\begin{bmatrix} 0 \\ 2 \end{bmatrix},
\begin{bmatrix} 0 \\ 3 \end{bmatrix},
\begin{bmatrix} 1 \\ 0 \end{bmatrix},
\begin{bmatrix} 1 \\ 1 \end{bmatrix},
\begin{bmatrix} 1 \\ 2 \end{bmatrix},
\begin{bmatrix} 1 \\ 3 \end{bmatrix}
\right\}
\]
Finally, we associate a reward signal $r_{t+1}$ with each state transition---defined by the tuple $\langle s_t, a_t, s_{t+1}\rangle$---to encompass (i) time into ventilation, (ii) physiological stability, i.e. whether vitals are steady and within expected ranges, (iii) failed SBTs or reintubation. The reward at each timestep is defined by a combination of sigmoid, piecewise-linear, and threshold functions that reward closely regulated vitals and successful extubation while penalizing adverse events: \looseness=-1
\vspace{-3mm}
\begin{flalign*}
&r_{t+1} = r^\textit{vitals}_{t+1} + r^\textit{vent off}_{t+1} + r^\textit{vent on}_{t+1} \text{, where} \\
& r^\textit{vitals}_{t+1} = C_{1} \sum\limits_{v} 
\left[\dfrac{1}{1 + e^{-(v_t - v_\textit{min})}} - \dfrac{1}{1 + e^{-(v_t - v_\textit{max})}} + \dfrac{1}{2} \right] \\
&  \left. \qquad\qquad\qquad \qquad \right. - C_{2}  \left[ \max\left(0, \dfrac{|v_{t+1} - v_t|}{v_{t}}- 0.2\right) \right], \\
& r^\textit{vent off}_{t+1} = \mathbbm{1}_{[s_{t+1}(\text{vent on}) = 0]}  \left[ C_3 \cdot \mathbbm{1}_{[s_t(\text{vent on}) = 1]} \right. \\ 
&   \left. \qquad + C_4\cdot \mathbbm{1}_{[s_t(\text{vent on}) = 0]} -  C_5\sum\limits_{v^\textit{ext}}\mathbbm{1}_{[v^\textit{ext}_t \,>\, v^\textit{ext}_\textit{max} \,||\, v^\textit{ext}_t \,<\, v^\textit{ext}_\textit{min}]} \right], \\
& r^\textit{vent on}_{t+1} = \mathbbm{1}_{[s_{t+1}(\text{vent on}) = 1]} \left[C_6\cdot\mathbbm{1}_{[s_{t}(\text{vent on}) = 1]}\right. \\ 
&\left. 
  \qquad\qquad\qquad\qquad\qquad\qquad\qquad\,\,\, - C_7 \cdot\mathbbm{1}_{[s_{t}(\text{vent on}) = 0]}\right].
\end{flalign*}
\vspace{-8mm}

\begin{figure}[ht!]
\centering{}
\captionsetup{justification=centering}
\begin{subfigure}{0.23\textwidth}
\centering
    \includegraphics[width=\textwidth]{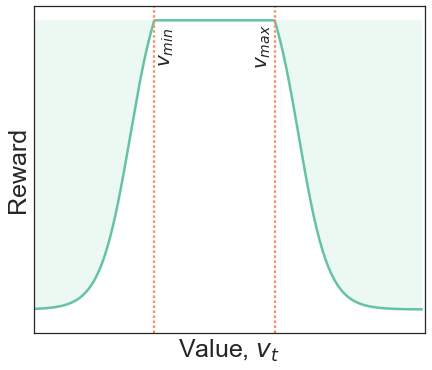}
    \caption{Exceeding threshold values}
    \label{fig:reward1}
\end{subfigure}
\begin{subfigure}{0.23\textwidth}
\centering
    \includegraphics[width=\textwidth]{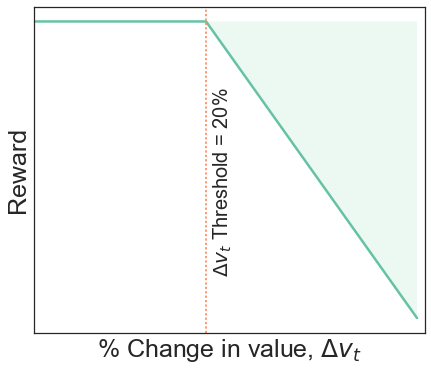}
    \caption{High fluctuation in values}
    \label{fig:reward2}
\end{subfigure}
	\vspace{1mm}
    \caption{Shape of reward from vitals, $r^\textit{vitals}_{t}(v_t)$}
    \vspace{-2mm}
\end{figure}

Here, values $v_t$ are the measurements of those vitals $v$ (included in the state representation $s_t$) believed to be indicative of physiological stability at time $t$, with desired ranges $[v_\textit{min}, v_\textit{max}]$. The penalty for exceeding these ranges at each time step is given by a truncated sigmoid function (Figure \ref{fig:reward1}). The system also receives negative feedback when consecutive measurements see a sharp change (Figure \ref{fig:reward2}). %\looseness=-1

Vital signs $v^\textit{ext}_t$ comprise the subset of parameters directly associated with readiness for extubation ($FiO_2$, $SpO_2$, and PEEP set) with weaning criteria defined by the ranges $[v^\textit{ext}_\textit{min}, v^\textit{ext}_\textit{max}]$. A fixed penalty is applied when these criteria are not met during extubation. The system also accumulates negative rewards for each additional hour spent on the ventilator, and a large positive reward at the time of successful extubation. Constants $C_1$ to $C_7$ determine the relative importance of these reward signals.

\subsection{Learning the Optimal Policy}
The majority of reinforcement learning algorithms are based on estimation of the \emph{Q}-function, that is, the expected value of state-action pairs $Q^\pi(s,a): S \times A \rightarrow \mathbb{R}$, to determine the optimal policy $\pi$. Of these, the most widely used is Q-learning, an off-policy reinforcement learning algorithm in which we start with an initial state and arbitrary approximation of the Q-function, and update this estimate using the reward from the next transition using the Bellman recursion for Q-values:

\vspace{2mm}
{\begin{nscenter}\small{$\hat{Q}(s_t, a_t) = \hat{Q}(s_t, a_t) + \alpha(r_{t+1} + \gamma\max\limits_{a \in \mathcal{A}}\hat{Q}_(s_{t+1}, a) - \hat{Q}(s_t, a_t))$} \end{nscenter}

where the learning rate $\alpha$ gives the relative weight of the current and previous estimate, and $\gamma$ is the discount factor.

Fitted Q-iteration (FQI) is a form of off-policy \textit{batch-mode} reinforcement learning that uses a set of one-step transition tuples:
\[
\mathcal{F} = \{(\langle s_t^n, a_t^n, s_{t+1}^n\rangle, r^n_{t+1}), n=1,...,|\mathcal{F}|\}
\]
to learn a sequence of function approximators $\hat{Q}_1, \hat{Q}_2...\hat{Q}_K$ of the value of state-action pairs, by iteratively solving supervised learning problems. Both FQI and Q-learning belong to the class of model-free reinforcement learning methods, which assumes no knowledge of the dynamics of the system. In the case of FQI, there are also no assumptions made on the ordering of tuples; these could correspond to a sequence of transitions from a single admission, or randomly ordered transitions from multiple histories. 
FQI is therefore more data-efficient, with the full set of samples used by the algorithm at every iteration, and hence typically converges much faster than Q-learning.  %\looseness=-1

The training set at the $k^{th}$ supervised learning problem is given by $\mathcal {TS} = \{(\langle s_t^n, a_t^n\rangle \,, \hat{Q}_k(s_t^n, a_t^n)) , {n=1,...,|\mathcal{F}|\}}$. As before, the Q-function is updated at each iteration according to the Bellman equation:
\[
\hat{Q}_k(s_t, a_t) \leftarrow r_{t+1} + \gamma\max\limits_{a \in \mathcal{A}}\hat{Q}_{k-1}(s_{t+1}, a) 
\]
where $\hat{Q}_1(s_t, a_t) = r_{t+1}$. The approximation of the optimal policy after $K$ iterations is then given by:
\[
\hat{\pi}^*(s) = \argmax\limits_{a \in \mathcal{A}}\hat{Q}_K(s,a)
\]
FQI guarantees convergence for many commonly used regressors, including kernel-based methods (\cite{ormoneit2002kernel}) and decision trees. In particular, extremely randomized trees (Extra-Trees: \cite{geurts2006extremely,ernst2005tree}), a tree-based ensemble method that extends on random forests by introducing randomness in the thresholds chosen at each split, has been applied in the past to learning large or continuous Q-functions in clinical settings (\cite{ernst2006clinical,EscandellMontero201447}). 

Neural Fitted-Q (NFQ, \cite{riedmiller2005neural}) on the other hand, looks to leverage the representational power of neural networks as regressors to fitted Q-iteration. \cite{nemati} use NFQ to learn optimal heparin dosages, mapping the patient hidden state to expected return. Neural networks hold an advantage over tree-based methods in iterative settings in that it is possible to simply update network weights at each iteration, rather than rebuilding the trees entirely.

\begin{algorithm}[ht]
  \SetAlgoLined
  %\textbf{function }{FQI$(\mathcal{F},\theta)$} \\
  \textbf{Input: \\} {One-step transitions $\mathcal{F} = \{\langle s_t^n, a_t^n, s_{t+1}^n\rangle, r_{t+1}^n\}_{n=1:|\mathcal{F}|}$; \\ 
  $\qquad $ Regression parameters $\theta$}; \\ Action space $\mathcal{A}$; subset size $N$ \\
  \textbf{Initialize} $Q_0(s_t,a_t) = 0 \quad \forall s_t \in \mathcal{F}, \, a_t \in \mathcal{A}$ \\
  \For{iteration $k = 1 \rightarrow K$}{
  {$\textit{subset}_N \sim \mathcal{F}$} \\
  {$S \leftarrow []$} \\
  \For{$i \in \textit{subset}_N$ }{
    {$Q_k(s_i, a_i) \leftarrow r_{i+1} + \gamma \max\limits_{a' \in \mathcal{A}} \left(\textit{predict}(\langle s_{i+1}, a'\rangle, \theta)\right)$}
    {$S \leftarrow \textit{append}(S, \langle(s_i, a_i), \, Q(s_i, a_i) \rangle)$}         
  }
  $\theta \leftarrow \textit{regress}(S)$
  % \If{converged}{break}
   }
  \KwResult{$\theta$}
  $\pi \leftarrow \textit{classify}(\langle s_t^n, a_t^n\rangle) $
   \caption{Fitted Q-iteration with sampling}
   \label{fqi}
\end{algorithm}

\section{Experimental Results}

After extracting relevant ventilation episodes from ICU admissions in the MIMIC III database (Section 3.1), and splitting these into training and test data, we obtain a total of 1,800 distinct admissions in our training set and 664 admissions in our test set. We interpolate time-varying vitals measurements using Gaussian processes or sample-and-hold interpolation, sampling at 10-minute intervals. This yields of the order of 1.5 million one-step transitions in the training set and 0.5 million in the test set respectively, where each transition is a 32-dimensional representation of patient state. \looseness=-1

As a baseline, we applied Q-learning to the training data to learn the mapping of continuous states to Q-values, with function approximation using a three-layer feedforward neural network. The network is trained using \emph{Adam}, an efficient stochastic gradient-based optimizer (\cite{kingma2014adam}), and $l_2$ regularization of weights. Each patient admission $k$ is treated as a distinct episode, with on the order of thousands of state transitions in each; the network weights are incrementally updated following each transition. Studying the change between successive episodes in the predicted Q-values for all state-action pairs in the training set (Figure \ref{qconv}), it is unclear whether the algorithm succeeds in converging within the 1,800 training episodes.
\begin{figure}[ht]
\centering
\captionsetup{justification=centering}
\includegraphics[width=0.48\textwidth]{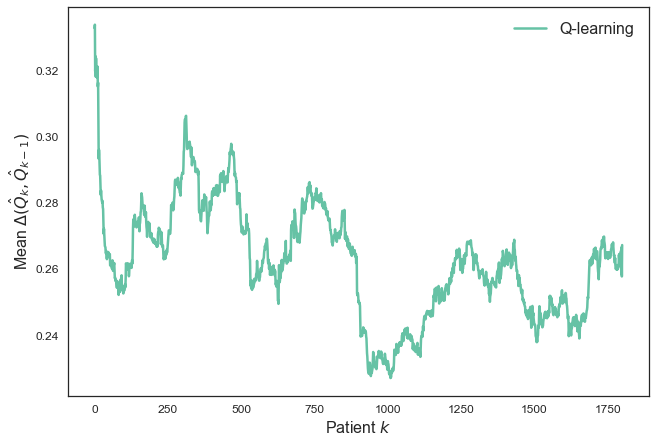}
\caption{Convergence of $\hat{Q}(s,a)$ using Q-learning.}
\label{qconv}
\end{figure} 

We then explored the use of FQI to learn our Q-function, first running with an Extra-Trees for function approximation. In our implementation, each iteration of FQI is performed on a random subset of 10\% of all transitions in the training set, as described in Algorithm \ref{fqi}, such that on average, each sample is seen in a tenth of all iterations. Though sampling increases the total number of iterations required for convergence, it yields significant speed-ups in building trees at each iteration, and hence in total training time. The ensemble regressor learns 50 trees, with regularization in the form of a minimum leaf node size of 20 samples. We present here results with FQI performed for a fixed number of 100 iterations, though it is possible to use a convergence criterion of the form $\Delta(Q_k, Q_{k-1}) \leq \varepsilon$ for early stopping, to speed up training further. %\looseness=-1

For comparison, we used he same methods to run FQI with neural networks (NFQ) in place of tree-based regression: we train a feedforward network with architecture and techniques identical to those applied in our function approximation for Q-learning. Convergence of the estimated Q-function for both regressors is measured by the mean change in the estimate $\hat{Q}$ for transitions in the training set (Figure \ref{fig:fqi_convergence}) which shows that the algorithm takes roughly 60 iterations to converge in both cases. However, NFQ yields approximately a four-fold gain in runtime speed, as expected, since with neural networks we can simply update weights rather than retraining fully at each iteration.

\begin{figure}[ht]
\centering
\captionsetup{justification=centering}
\includegraphics[width=0.48\textwidth]{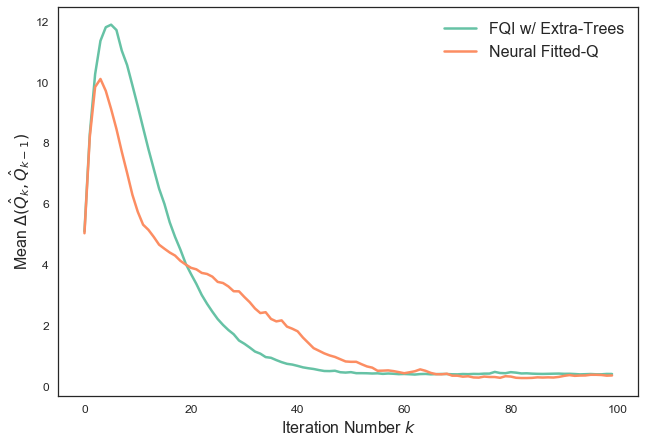} %fqi_conv11
\caption{Convergence of estimated $Q$ using FQI, given by the mean change in $\hat{Q}(s,a)$ over successive iterations.}
\label{fig:fqi_convergence}
\end{figure} 

\begin{figure*}[ht!]
\centering
\captionsetup{justification=centering}
\includegraphics[width=\textwidth]{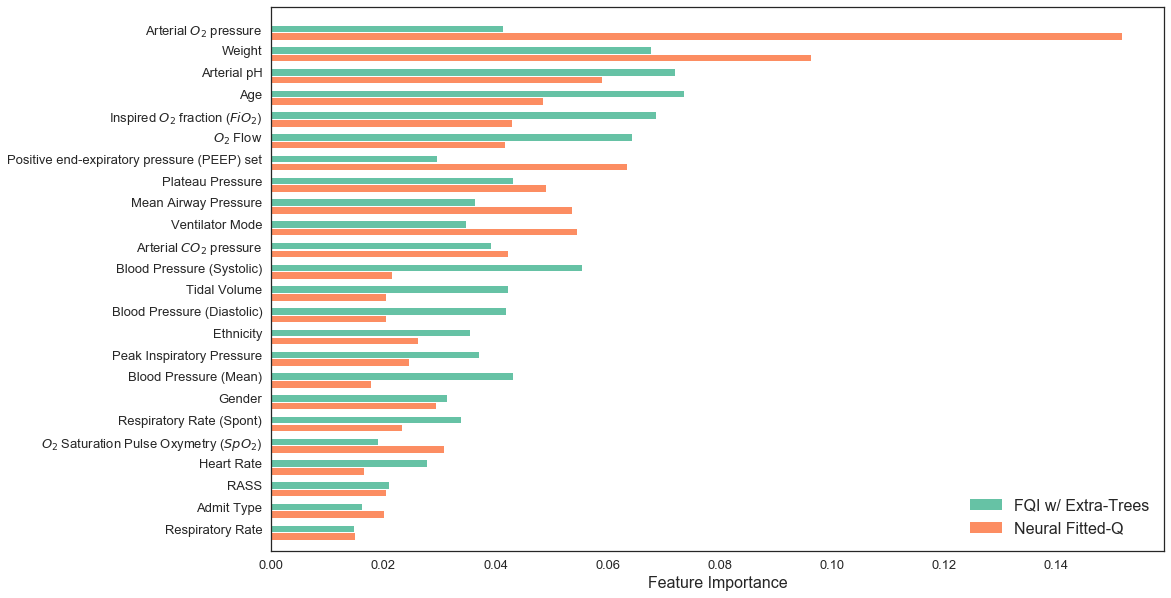}
\caption{Feature importances (from Gini importance scoring) for policies trained on optimal actions from FQIT \& NFQ. The relatively high weighting of indicators \emph{Arterial pH, $FiO_2$} and \emph{PEEP set} found is in agreement with typical protocol.}
\label{featImportances}
\end{figure*}

\begin{figure*}[ht]
\centering
\captionsetup{justification=centering}
\begin{subfigure}{0.48\textwidth}
\centering
    \includegraphics[width=\textwidth]{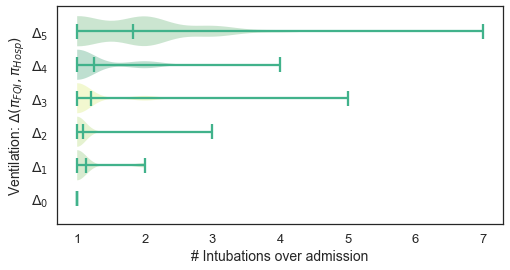}
    
        \vspace{-2mm}
    \caption{Ventilation Policy: Reintubations}
        \vspace{2mm}

    \label{fig:ventviolin1}
\end{subfigure}
\begin{subfigure}{0.48\textwidth}
\centering
    \includegraphics[width=\textwidth]{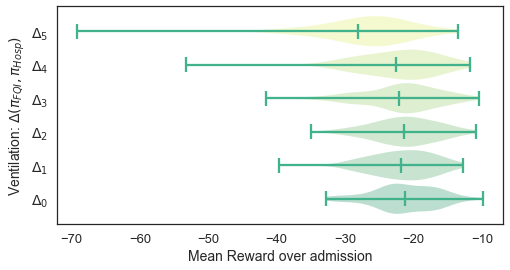}
    
        \vspace{-2mm}
    \caption{Ventilation Policy: Accumulated Reward}
        \vspace{2mm}
    \label{fig:ventviolin2}
\end{subfigure}
\begin{subfigure}{0.48\textwidth}
\centering
        \includegraphics[width=\textwidth]{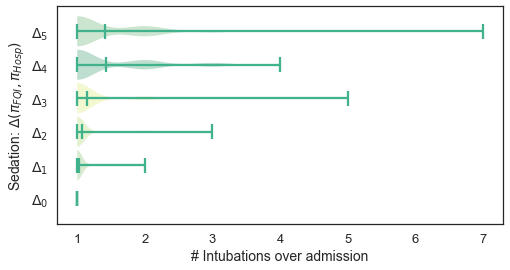}
        
                \vspace{-2mm}        
        \caption{Sedation Policy: Reintubations}        
        \vspace{2mm}

        \label{fig:sedviolin3}
    \end{subfigure}
\begin{subfigure}{0.48\textwidth}
\centering
        \includegraphics[width=\textwidth]{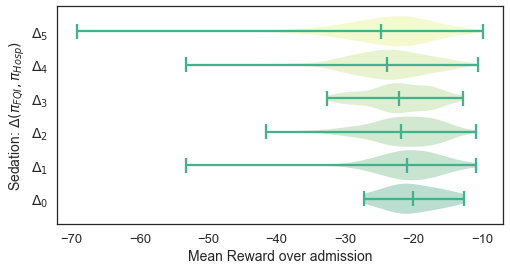}
        
                \vspace{-2mm}
        \caption{Sedation Policy: Accumulated Reward}
                \vspace{2mm}

        \label{fig:sedviolin4}
    \end{subfigure}
    %\vspace{-2mm}    
    \caption{Evaluating policy in terms of reward and number of reintubations suggests admissions where actions match our policy more closely are generally associated with better patient outcomes, both in terms of number of reintubations and accumulated reward, which reflects in part the regulation of vitals.}
    \label{violins}
\end{figure*}

The estimated Q-functions from FQI with Extra-Trees (FQIT) and from NFQ are then used to evaluate the optimal action, i.e. that which maximizes the value of the state-action pair, for each state in the training set. We can then train policy functions $\pi(s)$ mapping a given patient state to the corresponding optimal action $a \in \mathcal{A}$. To allow for clinical interpretation of the final policy, we choose to train an Extra-Trees classifier with an ensemble of 100 trees to represent the policy function. 

The relative importance assigned to the top 24 features in the state space for the policy trees learnt, when training on optimal actions from both FQIT and NFQ, show that
the five vitals ranking highest in importance across the two policies are arterial $O_2$ pressure, arterial pH, $FiO_2$, $O_2$ flow and PEEP set (Figure \ref{featImportances}). These are as expected---Arterial pH, $FiO_2$, and PEEP all feature in our preliminary HUP guidelines for extubation criteria, and there is considerable literature suggesting blood gases are an important indicator of readiness for weaning (\cite{Hoo1019}). On the other hand, oxygen saturation pulse oxymetry ($SpO_2$) which is also included in HUP's current extubation criteria, is fairly low in ranking. This may be because these measurements are highly correlated with other factors in the state space, such as arterial $O_2$ pressure (\cite{collins2015relating}), that account for its influence on weaning more directly. The limited importance assigned to heart rate and respiratory rate, which can serve as indicators of blood pressure and blood gases, 
are also likely to be explained by this dependence between vitals.

In terms of demographics, weight and age play a significant role in the weaning policy learnt: weight is likely to influence our sedation policy specifically, as dosages are typically adjusted for patient weight, while age is strongly correlated with a patient's speed of recovery, and hence the time necessary on ventilator support.

In order to evaluate the performance of the policies learnt, we compare the algorithm's recommendations against the true policy implemented by the hospital. Considering ventilation and sedation separately, the policies learnt with FQIT and  NFQ achieve similar accuracies in recommending ventilation (both matching the true policy in roughly 85\% of transitions), while FQIT far outperforms NFQ in the case of sedation policy (achieving 58\% accuracy compared with just 28\%, barely above random assignment of one of four dosage levels), perhaps due to overfitting of the neural network on this task. More data may be necessary to develop a meaningful sedation policy with NFQ. 

We therefore concentrate further analysis of policy recommendations to those produced by FQIT. We divide the 664 test admissions into six groups according to the fraction of FQI policy actions that differ from the hospital's policy: $\Delta_0$ comprises admissions in which the true and recommended policies agree perfectly, while those in $\Delta_5$ show the greatest deviation.
Plotting the distribution of the number of reintubations and the mean accumulated reward over patient admissions respectively, for all patients in each set (Figures \ref{fig:ventviolin1} and \ref{fig:ventviolin2}), we can see that those admissions in set $\Delta_0$ undergo no reintubation, and in general the average number of reintubations increases with deviation from the FQIT policy, with up to seven distinct intubations observed in admissions in $\Delta_5$. This effect is emphasised by the trend in mean rewards across the six admission groups, which serve primarily as an indicator of the regulation of vitals within desired ranges and whether certain criteria were met at extubation: mean reward over a set is highest (and the range lowest) for admissions in which the policies match exactly; mean reward decreases with increasing divergence of the two policies.
A less distinct but comparable pattern is seen when grouping admissions instead by similarity of the sedation policy to the true dosage levels administered by the hospital (Figures \ref{fig:sedviolin3} and \ref{fig:sedviolin4}). 

\section{Conclusion}
%\vspace{-1mm}
In this work, we propose a data-driven approach to the optimization of weaning from mechanical ventilation of patients in the ICU. We model patient admissions as Markov decision processes, developing novel representations of the problem state, action space, and reward function in this framework. Reinforcement learning with fitted Q-iteration using different regressors is then used to learn a simple ventilator weaning policy from examples in historical ICU data. We demonstrate that the algorithm is capable of extracting meaningful indicators for patient readiness and shows promise in recommending extubation time and sedation levels, on average outperforming clinical practice in terms of regulation of vitals and reintubations.
\looseness=-1

There are a number of challenges that must be overcome before these methods can be meaningfully implemented in a clinical setting, however: 
first, in order to generate robust treatment recommendations, it is important to ensure policy invariance to reward shaping: the current methods display considerable sensitivity to the relative weighting of various components of the feedback received after each transition. A more principled approach to the design of the reward function, for example by applying techniques in inverse reinforcement learning (\cite{ng2000algorithms}), can help tackle this sensitivity.
In addition, addressing the question of censoring in sub-optimal historical data and explicitly correcting for the bias that arises from the timing of interventions is crucial to fair evaluation of learnt policies, particularly where they deviate from the actions taken by the clinician. Finally, effective communication of the best action, expected reward, and the associated uncertainty, calls for a probabilistic approach to estimation of the Q-function, which can perhaps be addressed by pairing regressors such as Gaussian processes with Fitted Q-iteration.

Possible directions for future work also include increasing the sophistication of the state space, for example by handling long term effects more explicitly using second-order statistics of vitals, applying techniques in inverse reinforcement learning to feature engineering (as in \cite{levine2010feature}), or modeling the system as a partially observable MDP, in which observations map to some underlying state space. Extending the action space to include continuous dosages of specific drug types and settings such as ventilator modes and $FiO_2$ will also facilitate directly actionable policy recommendations. With further efforts to tackle these challenges, the reinforcement learning methods explored here will play a crucial role in helping to inform patient-specific decisions in critical care. %\looseness=-1

\bibliography{main}
\bibliographystyle{unsrtnat}

\end{document}